\documentclass[runningheads]{llncs}
\usepackage[T1]{fontenc}
\usepackage{graphicx}
\begin{document}
\title{Investigating the changes in BOLD responses during viewing of images with varied complexity: An fMRI time-series based analysis on human vision
}
%
%
\author{Naveen Kanigiri \inst{1} \and
Manohar Suggula\inst{1} \and
Debanjali Bhattacharya\inst{1} \and 
Neelam Sinha \inst{1} }
\authorrunning{Naveen Kanigiri et al.}
%
\institute{International Institute of Information Technology, Bangalore\\
\email{debanjali.bhattacharya@iiitb.ac.in}
}
\maketitle              
\begin{abstract}
Functional MRI (fMRI) is widely used to examine brain functionality by detecting alteration in oxygenated blood flow that arises with brain activity. This work aims to investigate the neurological variation of human brain responses during viewing of images with varied complexity using fMRI time series (TS) analysis. Publicly available BOLD5000 dataset is used for this purpose which contains fMRI scans while viewing 5254 distinct images of diverse categories, drawn from three standard computer vision datasets: COCO, Imagenet and SUN. To understand vision, it is important to study how brain functions while looking at images of diverse complexities. Our first study employs classical machine learning and deep learning strategies to classify image complexity-specific fMRI TS, represents instances when images from COCO, Imagenet and SUN datasets are seen. The implementation of this classification across visual datasets holds great significance, as it provides valuable insights into the fluctuations in BOLD signals when perceiving images of varying complexities. 
Subsequently, temporal semantic segmentation is also performed on whole fMRI TS to segment these time instances. 
The obtained result of this analysis has established a baseline in studying how differently human brain functions while looking into images of diverse complexities. Therefore, accurate identification and distinguishing of variations in BOLD signals  from fMRI TS data serves as a critical initial step in vision studies, providing insightful explanations for how static images with diverse complexities are perceived.

\end{abstract}
\section{Introduction}
\label{sec:intro}

Functional Magnetic Resonance Imaging (fMRI) is a widely utilized tool in neuroscience research for analyzing brain functional connectivity during specific tasks or resting-state conditions. Over the past decade, the field of machine vision has undergone a transformative evolution with the introduction of large-scale computer vision image datasets and advanced statistical learning techniques. These resources have enabled various computer vision tasks such as object detection, localization, segmentation, and classification. However, the exploration of human visual perception has remained limited due to the complex experimental procedures required to generate a sufficient number of high-quality fMRI neuro-image data that represents vision. The recent release of the publicly available dataset called "BOLD5000" \cite{b1} has opened a new direction for studying the dynamics of the human brain during visual tasks with greater details. The dataset comprises fMRI scans acquired from subjects while viewing 5000 images, selected from three renowned computer vision datasets: COCO, Imagenet, and SUN. These images encompass diverse context and complexities that mainly facilitate the studies of various computer vision-related tasks like object detection, localization, segmentation and classification.
There are few studies reported in literature that used BOLD5000 dataset for tasks like classification of well-separable image classes \cite{b2}, for pre-training to predict cognitive fatigue in traumatic brain injury \cite{b3} and for neural encoding\cite{b4}. \par 
Different from these studies, in our work, we have utilized the BOLD fMRI time series (TS) to categorize images with diverse complexities across these visual datasets. While there exist several literature on TS classification using machine learning (ML) and deep learning  (DL) models \cite{b5,b6,b7,b8,b9,b10,b13,b15}, the exploration of fMRI TS data that accurately represents human vision is limited. This motivates us to study the potential utilization of fMRI TS data in categorizing images across visual datasets. By doing so, we aim to uncover the unique functioning of the human brain when exposed to images with diverse context and complexities. 
The execution of this classification process across visual datasets using fMRI TS data is of significant importance, as it sheds light on the differences in BOLD signals during the perception of images with diverse complexities. Therefore, predicting these differences from fMRI TS serves as a crucial initial step in vision studies, offering meaningful explanations for the perception of static images.
The contributions of this paper are:
\begin{itemize}
    \item Categorization of images across distinct visual datasets using fMRI TS, as derived from each active voxel during visual tasks.
\end{itemize}
\begin{itemize}
    \item Temporal semantic segmentation to label each time instance of whole fMRI TS according to image complexity- representing distinct datasets.
\end{itemize}

\begin{figure}[htb]
\centering
\begin{minipage}[b]{0.3\linewidth}
  \centering
  \centerline{\includegraphics[width=3.55cm]{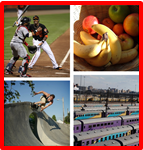}}
  \centerline{(a)}\medskip
\end{minipage}
\begin{minipage}[b]{0.3\linewidth}
  \centering
  \centerline{\includegraphics[width=3.6cm]{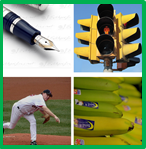}}
  \centerline{(b)}\medskip
\end{minipage}
\begin{minipage}[b]{0.32\linewidth}
  \centering
  \centerline{\includegraphics[width=3.67cm]{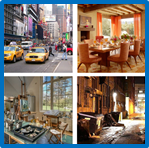}}
  \centerline{(c)}\medskip
\end{minipage}
\caption{Sample images, taken from the three computer vision datasets having different complexities: (a) COCO (\textit{Red box}): contains multiple objects and actions, (b) Imagenet (\textit{Green box}): contains single-focused object and (c) SUN (\textit{Blue box}): contains indoor and outdoor scenes.}
\label{fig:imgs}
\end{figure}

\begin{figure}[htp]
    \centering
    \includegraphics[width=12cm]{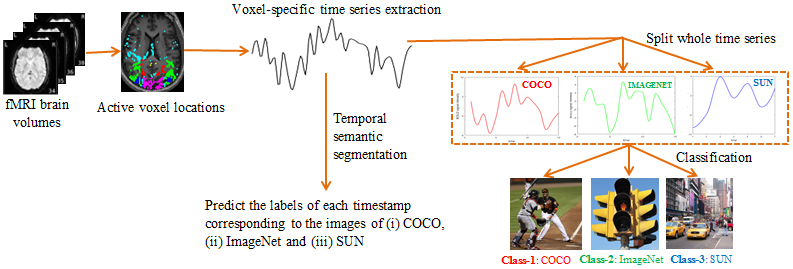} 
    \caption{Block schematic of the proposed methodology}
    \label{fig:FlowChart}
\end{figure}

\section{Dataset}

The publicly available BOLD5000 dataset is used in this work \cite{b1} that contain human fMRI scans of 4 participants while viewing a total of 5254 images of different categories in random order. These images were drawn from three standard computer vision datasets having diverse image complexities: (i) 2000 images of multiple indoor/outdoor objects interacting with each other, taken from \textit{Common Objects in Context (COCO)} database, (ii) 1916 images of in-focus singular object, drawn from \textit{Imagenet} database and (iii) 1338 scene images of \textit{Scene Understanding (SUN)} database, which were more scenic, with less emphasis on any particular object or action. Figure \ref{fig:imgs} shows some sample images of these three datasets. Each functional session consisted of 9 to 10 trails where in each trail 37 stimuli (images) were presented randomly to the participants. Each fMRI sessions was roughly 1.5 hours long for all subjects. The fMRI data was acquired using a 3T Siemens Verio MR scanner using a 32-channel phased array head coil. Further details on subject demographics, stimuli selection, fMRI scan acquisition and data pre-processing procedures can be found in \cite{b1}. 

\section{Proposed Methodology: Complexity-based image categorization across visual datasets and temporal semantic segmentation}

The block diagram of the proposed methodology is shown in the Figure \ref{fig:FlowChart}. The whole TS is extracted from each active voxel in each fMRI trail. For categorizing the TS of images across visual datasets (Imagenet, COCO and SUN), each of the whole TS is split into three parts according to the images of the three datasets as shown to the participants during each fMRI trail. These voxel-specific TS are then used to train ML and DL models for image categorization into corresponding datasets. For temporal semantic segmentation (described in Section \ref{ref: tss}), DL framework is applied on the whole TS, for predicting the labels of each timestamp whether corresponds to the images of COCO, Imagenet or SUN, as seen by the participants. The code of this experiment will be available in the GitHub link: 
https://github.com/Naveen7102/FMRI-Time-Series-Classification

\subsection{FMRI TS Extraction}
\label{ref:FMRI Time Series Extraction}

\begin{figure}[htp]
\centering
\includegraphics[width=8.5cm]{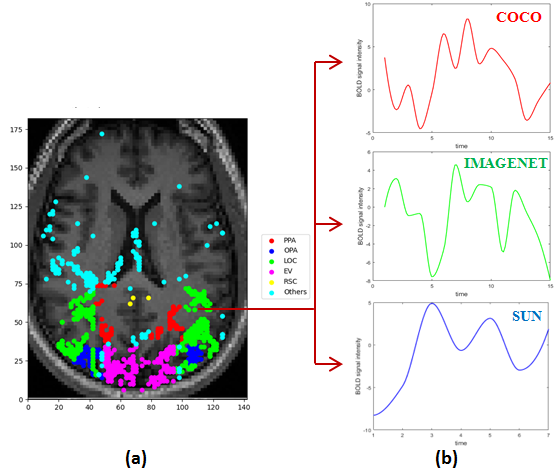} 
\caption{fMRI TS extraction from active voxel location in brain. \textbf{(a)} Mapping of the location of voxel activation on fMRI image for a representative subject. Voxel activation at different visual ROIs (as defined in \cite{b1}) are shown by different colors. \textbf{(b)} it shows the extracted TS that represents 3 distinct datasets for a specific active voxel (top: COCO, middle: Imagenet, bottom: SUN).}
\label{fig:TimeSeries}
\end{figure}

In order to extract BOLD fMRI TS, it is necessary to have the information about active voxels locations in brain while viewing the images. In our study, statistical parametric mapping (SPM) toolbox \cite{b17} is utilized to get active voxel locations from 4D \textit{(x,y,z,t)} fMRI data. The locations of active voxels are mostly found within 5 visual region of interests (ROIs) as defined in \cite{b1}, which are the parahippocampal place area (PPA), the retrosplenial complex (RSC), the occipital place area (OPA), Early visual area (EV) and lateral occipital complex (LOC). Few voxels which are found to be active in other sub-cortical regions, labeled as "others" in this study (Figure~\ref{fig:TimeSeries}: \textit{left}). 
From each fMRI trail and for each active voxel, the whole TS which is the representation of BOLD intensity distribution over time, was extracted. Further, in order to obtain the image complexity-specific fMRI TS, the whole TS of length 37 (since, the no. of stimuli = 37, at each trial) is split into three image complexity-specific TS as shown in Figure~\ref{fig:TimeSeries} (\textit{right}). Each of these three TS illustrates the BOLD intensity distribution while viewing images from COCO (\textit{Red}), Imagenet (\textit{Green}) and SUN (\textit{Blue}). It is to be noted that, the length of these three sets of image complexity-specific TS are not same due to randomness in total number of image presentation to the participants from these three datasets. The obtained TS are detrended and Z-score normalized before it is fed to ML/DL model.

\subsection{Model Architecture for TS Classification}
\label{ref: model architecture for classification}

Various ML classifiers and neural network architectural choices for fMRI TS classification are described here. SVM with RBF kernel, AdaBoost and XGBoost are used as ML classifiers in order to categorize the image complexity-specific TS of (i) COCO, (ii) Imagenet and (iii) SUN. Subsequently, Long Short-Term Memory (LSTM) and Bi-directional LSTM (Bi-LSTM) neural network models are also used for TS classification. The extracted image complexity-specific TS of each active voxel are fed as the input to the classifies.
The architecture of LSTM (and Bi-LSTM) is shown in Figure~\ref{fig:models}(a). In both models, two LSTM and Bi-LSTM layers are used. The input TS is fed into first dense layer of size 32 units. The first LSTM, having size 64 units (for Bi-LSTM, it is 128 units) is then applied on this dense layer, followed by a dropout layer (dropout value 0.5) in order to prevent overfitting. Similar LSTM with 32 units (for Bi-LSTM, it is 64 units) and drop out layer are added further and flattened which is followed by two dense layers of size 64 units and 32 units, respectively. The fourth dense layer of size 3 units is added at the end for final prediction. 'ReLU' activation function is used in all dense layers except the last layer where 'softmax' (for 3-class classification)/'sigmoid' (for binary classification) activation function is applied for final prediction.

\subsection{Temporal Semantic Segmentation}
\label{ref: tss}
Semantic segmentation is defined as classifying a specific class of data and segregating it from the rest of the data classes by overlaying it with a segmentation mask. While semantic segmentation is very common in case of image data; where the task is to label each pixel of an image according to its class category, no works have been reported on 1D semantic segmentation of fMRI TS, representing brain activity over time. In this work, the idea of 2D semantic segmentation is applied on 1D TS data. The goal is to classify the BOLD signal intensity at each timestamp of the whole TS with a corresponding image complexity-specific class label (\textit{'L'}) of what it belongs to (i.e. whether it is COCO (\textit{L=1}) or Imagenet (\textit{L=2}) or SUN (\textit{L=3})). \par
\begin{figure}[htp]
\centering
\includegraphics[width=10cm]{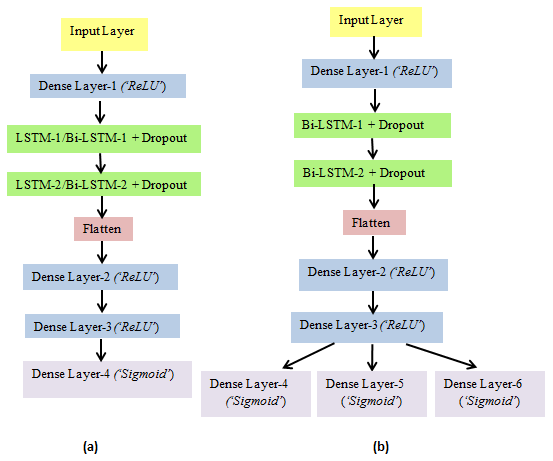}  
\caption{\textbf{(a)} LSTM/Bi-LSTM model architecture as considered in this study. \textbf{(b)} Bi-LSTM architecture for temporal semantic segmentation of fMRI TS}
\label{fig:models}
\end{figure}

Bi-LSTM architecture as shown in Figure~\ref{fig:models}(b) is used for temporal semantic segmentation. Here, the whole TS of length \textit{t= 37} is fed to the first dense layer, consisting of 64 units followed by a series of two Bi-LSTM layers of size 256 units and 512 units respectively, with dropout layers having dropout value 0.5. Two dense layers of size 128 units and 64 units are added further. Similar to classification framework, the choice of activation function for these dense layers are 'ReLU'. Towards end of this model, three dense layers are added with 37 units in each, for the purpose of segmenting the whole TS into image complexity-specific class. For temporal semantic segmentation, 'sigmoid' activation function is used in the last dense layer of the Bi-LSTM network model.

\subsection{Training}
\label{ref:training}

In both LSTM and Bi-LSTM network, binary cross-entropy loss with Adam optimizer is used for training the TS of distinct image categories. Learning rate is set to 0.001 with batch size of 20. Dropout regularization technique is applied to enhance the performance. The learning rate is set to 0.1 in case of ML classifiers- AdaBoost and XGBoost. In both ML and DL models, 10-fold cross validation is used. The input TS data are randomly split into train, test and validation, where 80\% data is used for training, 10\% data is used for validation and rest 10\% data is used to test the model. No data leakage is allowed between three splits. 
Due to variable length of the TS of COCO, Imagenet and SUN, zero is appended to make the length of each TS equal with the length of whole TS. Moreover, in order to ensure that the models are not predicting based on the number of zeros, the same TS sequence is repeated, instead of zero padding.
Kaggle notebooks and Keras are used to conduct the entire experiment. Kaggle provides 16GB of NVIDIA Tesla P100 GPU.

\section{Experimental Result and Analysis}

In this paper, the authors examine the ability of ML and DL models to categorize fMRI TS of varying image complexities across visual datasets of COCO, Imagenet and SUN. 

\begin{figure}[htp]
\centering
\includegraphics[width=12.5cm]{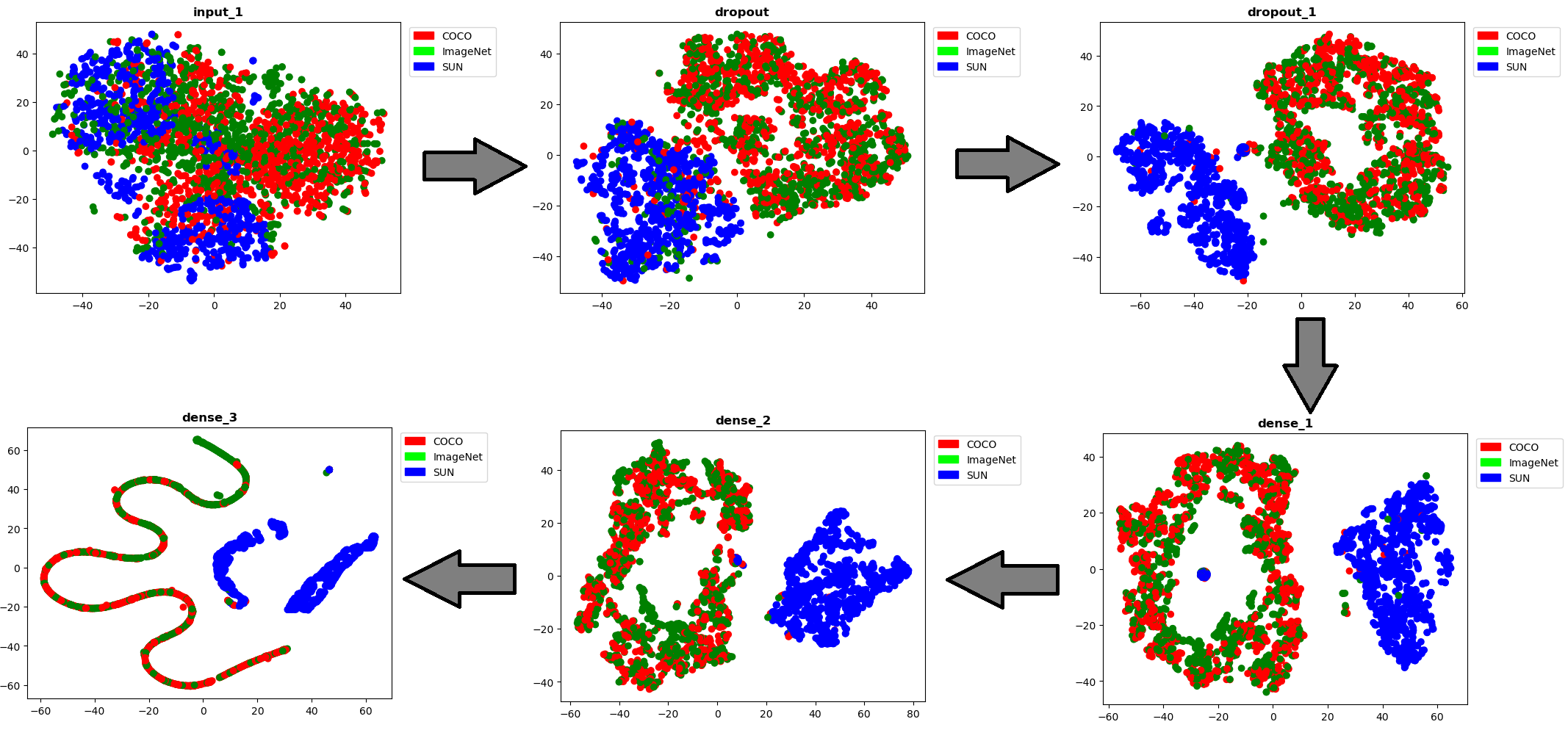}  
\caption{t-SNE visualization for Bi-LSTM classification across visual datasets is shown for input fMRI time-series (\textit{top row-left}) Bi-LSTM layer-1, Bi-LSTM layer-2, Dense Layer-1, Dense Layer-2 and softmax layer (\textit{bottom row-left}); that reveals well separability of BOLD signals of images between SUN, Imagenet and COCO datasets. However, classification for the pair Imagenet Vs. COCO yields less accuracy due to similarity in spatial context among 40\% of total images which were presented to the subjects from these two databases during fMRI sessions.}
\label{fig:tsne1}
\end{figure}

\begin{table*}[h]
\caption{Mean test accuracy for binary and ternary classification using different machine learning and deep learning models}
\label{table:classification}
\centering
\begin{tabular}{{|c|c|c|c|c|c|}}
\hline
 &Model & 3-class  &Imagenet   &Imagenet   &COCO   \\
 & &classification  &Vs. SUN  &Vs.COCO  &Vs. SUN  \\
 \hline
Machine &SVM(rbf) &0.74  &0.92  &0.64  &0.93 \\
learning &AdaBoost &0.67 &0.98 &0.63 &0.98 \\
& XGBoost &\textbf{0.76} &\textbf{0.99} &0.65 &\textbf{0.99} \\
\hline
Deep &LSTM &0.75 &0.98 &0.63 &0.985 \\
learning &Bi-LSTM &\textbf{0.76} &\textbf{0.99} &0.64 &\textbf{0.99} \\
\hline
\end{tabular}
\end{table*}

\subsection{fMRI TS Classification}

Different ML and DL models like SVM (with RBF kernel), AdaBoost, XGBoost and LSTM, Bi-LSTM are used to train the image complexity-specific fMRI TS of COCO, Imagenet and SUN. Using 10-fold cross-validation, the highest test accuracy of 76\% is obtained for 3-class classification using Bi-LSTM and XGBoost across all subjects. The result is improved drastically when 2-class classification is performed between fMRI TS of image complexities of two datasets, taken pairwise. As seen in Table~\ref{table:classification}, the cases of SUN Vs. Imagenet and SUN Vs. COCO yield the best performance with 99\% classification accuracy. This could be attributed to the fact that the images contain in SUN dataset is very distinct from images contain in COCO and Imagenet. However, the case of Imagenet Vs. COCO, yields only 65\% accuracy, which could be due to the similarities in the images, such as objects placed in similar backgrounds. It is observed that nearly 40\% of total images which were presented to participants from COCO and Imagenet, have high similarity in spatial context. One example of such spatial context similarity is shown in Figure~\ref{fig:imgs}: the baseball ground image in COCO and Imagenet that contain similar information, leading to high probability of misclassification which requires further investigation.
The classification performance is further explained by t-distributed stochastic neighbor embedding (t-SNE) visualization which is shown in Figure~\ref{fig:tsne1}. As seen from this figure, t-SNE visualization of temporal features of Bi-LSTM reveal clear differences in the BOLD TS among these datasets.  
This is to be noted that no significant changes in classification accuracy is found when TS is appended by the same signal values instead of zero-padding (as described in Section~\ref{ref:training}). Table~\ref{table:classification} shows the mean classification accuracy, as obtained across all subjects.

\subsection{Temporal Semantic Segmentation of Whole TS}

\begin{figure*}[htp]
    \centering
    \includegraphics[width=12.5cm]{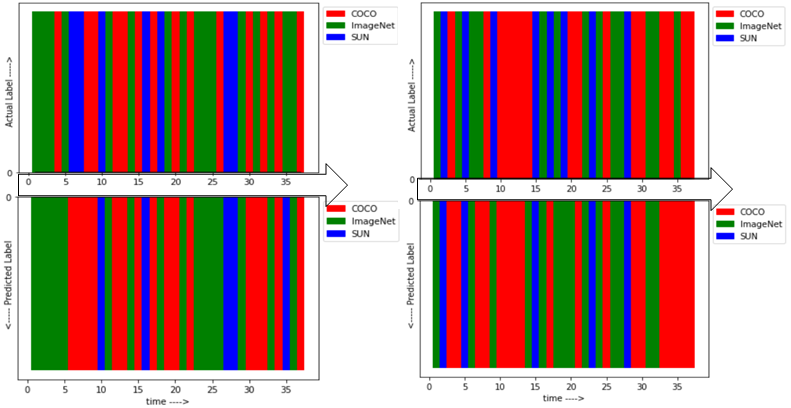}
    \caption{Ribbon plot for temporal semantic segmentation of whole fMRI TS. The labels for COCO, Imagenet and SUN are shown by Red, Green and Blue color respectively. In both plots, the actual label and predicted label is shown above and below time axis (Black arrow) respectively. The differences in color between actual and predicted label points the instances of misclassification.}
    \label{fig:SegmentationResults}
\end{figure*}

For temporal semantic segmentation, Bi-LSTM network model is chosen since it showed the best performance in fMRI TS classification. The whole TS of length 37 is used to train Bi-LSTM to classify BOLD intensity of each timestamp, according to the specific image category it represents. This results to semantic segmentation of the entire TS in 1D space. The output of temporal semantic segmentation is shown by Ribbon plot in Figure~\ref{fig:SegmentationResults}. The labels of each timestamp of the input TS and the predicted TS are color coded and plotted above and below the time axis respectively, in order to better visualize the performance of temporal semantic segmentation in 1D space. The differences in color between actual and predicted label in Figure~\ref{fig:SegmentationResults} points the instances of misclassification. The highest Dice coefficient is found to be 0.83, 0.77 and 0.6 in segmenting the TS represents COCO, Imagenet and SUN respectively. However, the average Dice coefficient across all subjects is decreased to 0.5, which requires further investigation.\par
To the best of our knowledge, this is the first study that utlized BOLD5000 dataset and fMRI TS to categorize images having diverse complexities into corresponding visual datasets of Imagenet, COCO and SUN. The dissimilarities in visual information conveyed by these datasets are apparent. For example, Imagenet images primarily feature unoccluded objects that are centrally focused, occupying a significant portion of the image and exhibiting uniform illumination. On the other hand, COCO images depict single or multiple objects commonly found in everyday life, appearing at different scales. SUN dataset consists of images capturing natural environments with backgrounds that are cluttered, varying illumination, and occlusions, without specific emphasis on any particular object(s). These distinctions have been well examined in recent literature \cite{b17,b18}. In this study, we have delineated these distinctions by classifying fMRI TS data of visual stimuli into their respective visual datasets, aiming to understand how differently human brain functions while looking at images having different complexities. The proposed methodology outperforms previous work by Jamalian et.al. \cite{b2} on BOLD5000 dataset in which authors used sequence models to classify only three well-separable classes of images: animal, artifact and scenes and achieved the accuracy of 68\%. There are two more studies reported in literature that used BOLD5000 dataset for other tasks. Jaiswal et.al. \cite{b3} used BOLD5000 images to pretrain deep neural network models for predicting cognitive fatigue in traumatic brain injury. Oota et.al. \cite{b4} used BOLD5000 dataset to study brain encoding models that aims to reconstruct fMRI brain activity given a stimulus. Contrary to these studies, the current work presented a different approach that focused on analysing the efficacy of utilizing fMRI TS to categorize image complexities across different natural image datasets.  

\section{Conclusion}

In the present study, we have performed fMRI TS analysis to classify images of varying complexity into distinct datasets of COCO, Imagenet, and SUN. Visualization of temporal features through t-SNE of Bi-LSTM features revealed clear differences in the BOLD TS among these datasets. This differentiation facilitated successful classification of the BOLD TS into their respective datasets. There are few limitations of the current study. Firstly, the study did not explore how these stimuli can influence network responses. Thus, in future, whether stimuli with different contexts and complexities induce unique network connectivity patterns needs to be investigated.
The second limitation of this study is the small sample size of only four subjects in the BOLD5000 dataset, which hinders drawing conclusive conclusions. To improve the generalizability of the study, it is essential to include more participants, additional fMRI sessions, and a wider variety of stimulated images. While the inclusion of 5,254 images of varying complexity is substantial for studying brain visual dynamics using fMRI, it remains relatively small as compared to the rich visual experiences encountered in our everyday life.
Nevertheless, as a baseline work, the results showed a good foundation for future fMRI research on how the brain represents vision. 

\subsubsection{Acknowledgements} 
The authors would like to thank Mphasis F1 Foundation, Cognitive Computing grant to conduct research at IIIT Bangalore.


%
%
%
%

\end{document}